\documentclass[conference]{IEEEtran}
\usepackage{newtxtext}
\usepackage{booktabs}
\usepackage{graphicx}
\usepackage{amsmath}
\usepackage{xcolor}
\definecolor{linknavy}{HTML}{1A3D7C}
\usepackage[colorlinks=true, linkcolor=linknavy, citecolor=linknavy,
            urlcolor=linknavy, breaklinks=true]{hyperref}

\makeatletter
\renewenvironment{abstract}{%
  \normalfont\footnotesize\noindent\textbf{\textit{Abstract.}}\ \ignorespaces}%
  {\par\vskip 0.5\baselineskip}
\renewenvironment{IEEEkeywords}{%
  \normalfont\footnotesize\noindent\textbf{\textit{Index Terms.}}\ \ignorespaces}%
  {\par}
\makeatother

\title{Impact Is Not Invalidation:\\
Ask About the Claim, Not the Diff}

\newif\ifanon\anonfalse

\ifanon
  \author{\IEEEauthorblockN{Anonymous Author(s)}
  \IEEEauthorblockA{Submission under double-blind review}}
\else
  \author{\IEEEauthorblockN{Atul Anand}
  \IEEEauthorblockA{\textit{Thomson Reuters}\\
  Bangalore, India}}
\fi

\begin{document}
\maketitle

\begin{abstract}
Memory systems for coding agents must decide, when a repository changes,
which of their stored claims have become false. Content anchoring
invalidates a claim whenever the artifact it came from changes, which
fires constantly. Semantic-equivalence classification asks whether a diff
preserves behavior, a question about the diff rather than about any stored
claim. We show the second signal fails for a reason unrelated to model
capability: asked whether a commit preserves behavior, five models
spanning a 40$\times$ price range fire on 59--72\% of real commits and
reach precisions of only 0.291 to 0.329 against a 0.25 base rate. Asked
instead whether one specific claim still holds, the same models on the
same diffs reach 0.705 to 0.974. A control that hands the
behavior-preservation judge the claim text, changing only the question,
moves precision by 0.010 and 0.016; changing the question moves it by 0.49
and 0.65. We also compare against \texttt{pytest-testmon}, a deployed
regression-test selector with coverage-derived dependency data: it reaches
0.868 recall at 0.415 precision, so near-complete knowledge of what a
change can reach does not identify what it falsifies. Ground truth is
execution, not annotation: a claim is a test function passing at commit
$t$, and it has flipped if that same assertion text fails at $t+1$.
Building this required an observation we did not find in prior work. On a
CI-gated mainline a commit that leaves a pre-existing test failing cannot
merge, so the naive construction has an empty positive class by design. We
report 10{,}369 claims with 184 execution-verified flips mined from 23
Python libraries, splits held out by repository, a post-knowledge-cutoff
split, a shuffled-diff null, a paraphrase control, and a
leave-one-repository-out analysis over 17 repositories.
\end{abstract}


\section{Introduction}

A coding agent that remembers things about a repository has to decide when
to forget them. If it recorded that \texttt{parse\_url} returns a tuple,
and someone changed it to return a dataclass, the memory is now wrong.
Serving it produces a confident error that is expensive to trace.

The literature offers two ways to make that decision, and they come from
different communities.

Work on memory and staleness has established that precision is the metric
that matters. A false ``this is superseded'' verdict destroys information
and cannot be recovered, while a false negative can be corrected later.
Systems in this line anchor a claim to the content it came from and
withdraw the claim when that content changes~\cite{eagraph}. This has
recall near one and poor precision, because most changes to a file do not
affect most claims about it.

Work on software semantics has execution oracles and asks whether a change
preserves behavior~\cite{changeguard,semadiff}. This is a more selective
signal. It is also a question about the diff, not about any stored claim,
and those are different questions. A pure rename preserves behavior while
making every claim that names the old symbol false. An internal
optimization changes the call sequence while leaving a claim about the
module's purpose true.

This paper asks whether the difference between those two questions is
worth anything in practice, and finds that it is worth most of the
available accuracy.

\begin{figure*}[t]\centering
\includegraphics[width=0.99\textwidth]{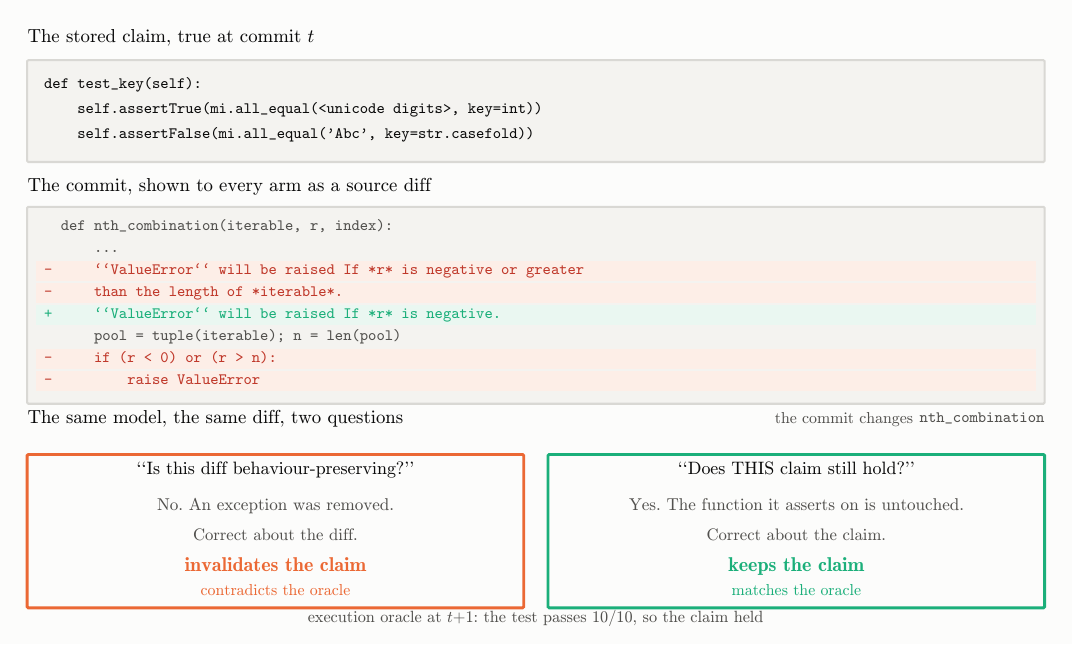}
\caption{A real case from our benchmark. The commit removes an argument
check from \texttt{nth\_combination}, so it does change behaviour and a
semantic-equivalence judge is right to say so. The stored claim is about
\texttt{all\_equal}, which the commit never touches, and execution
confirms it still holds. Answering ``did behaviour change'' correctly is
not the same as answering ``is this claim still true''.}
\label{fig:motivating}
\end{figure*}

Figure~\ref{fig:motivating} makes the distinction concrete, and shows that
the failure is not a matter of the judge being wrong about the code.

\noindent\textbf{Contributions.}
\begin{enumerate}
\item An execution-grounded benchmark for claim invalidation: 10{,}369
claims mined from 23 Python libraries with 184 flips verified by running
tests, not by annotation.
\item A construction result. The obvious way to build such a benchmark
produces no positive examples on any maintained repository, because CI
blocks the commits that would create them. We give a construction that
works and explain why.
\item A measurement that separates the question from the evidence. Giving
a behavior-preservation judge the claim text changes almost nothing;
changing the question changes almost everything.
\item Evidence that this is a property of the question rather than of
model scale, across five models spanning two providers and roughly a
40$\times$ price range.
\end{enumerate}

\section{Problem}

Let a claim $c$ be a statement about a repository that was true at commit
$t$. Let $d$ be the diff from $t$ to $t+1$. Claim-relative invalidation
predicts whether $c$ is false at $t+1$.

This differs from the two signals in use.

Content anchoring computes a digest over the artifact $c$ was derived from
and invalidates when the digest changes. It does not consult $c$.

Semantic-equivalence classification predicts whether $d$ changes
externally observable behavior. It also does not consult $c$. A commit
that changes behavior somewhere invalidates nothing in particular; a
commit that preserves behavior can still falsify a claim about a name.

The predicate we want is $\Pr[c \text{ false at } t+1 \mid c, d]$, which
conditions on both.

\noindent\textbf{Why not just run the tests.}
Execution decides this exactly, and a reviewer will raise it first. A
deployed store holds thousands of claims and commits arrive continuously,
so re-verifying everything costs $O(|\text{store}|)$ test runs per commit.
That is the cost memory systems exist to avoid. Execution is the
evaluation oracle here, not a deployable method. The question is how close
an $O(1)$-per-claim textual predictor gets.

\section{Benchmark}

\subsection{Claims and oracle}

A claim is a test function together with its assertions, passing
deterministically at commit $t$. It has flipped if the same assertion text
fails at $t+1$. There is no human judgment and no judge model.

We run every candidate $N=10$ times at $t$ and discard any claim that is
not deterministically passing, then $N=10$ times at $t+1$ and discard any
claim whose post-change outcome is not deterministic. Tests that produce
no verdict at all, which happens when a collection error aborts a pytest
batch, are recorded as unknown and discarded rather than counted as
failures. The discard rate is 0.16\%.

\subsection{CI-gating makes the naive construction empty}

The natural construction is to find tests that pass at $t$, are not
modified by the commit, and fail at $t+1$. We mined 1{,}005 such claims
and found zero flips.

This is not a sampling accident. A commit that leaves a pre-existing test
failing does not pass CI, so it does not reach mainline. On any repository
with CI, the population of unmodified tests that break is empty by
construction. Changing repositories does not help.

The flips that exist are in commits where the author changed behavior and
updated the affected test in the same commit. The test edit is the
evidence that a claim flipped. So we evaluate the parent's assertion text
against the child's source: check out $t+1$, restore the parent's version
of every test file the commit touched, and run. What executes at $t+1$ is
exactly the claim that was true at $t$.

Because the restored assertion also appears in the commit's test diff, the
model is shown a source-only diff with test files excluded. We verified
that no test content reaches the prompt.

\subsection{Mining}

Flips are rare, so uniform commit sampling wastes almost all of its
budget. We rank commits by the number of assertion lines deleted from test
files, which is the direct signature of a retired expectation. Ranking by
commit-message keywords (``fix'', ``break'', ``remove'') does worse: the
first flip commit we verified by hand is titled \emph{``funcutils: forward
defaulted and kw-only args as keywords''} and matches none of them, while
many ``fix'' commits touch only documentation. Assertion-deletion ranking
raised flip density from 0.36\% to 2.0\%.

\subsection{Corpus}

Twenty-three Python libraries were mined, of which ten yielded flips:
187 commits, 10{,}369 claims, 184 flips (1.8\%). Table~\ref{tab:corpus}
gives the breakdown, including the repositories that produced no
positives, since the rate at which mining comes up empty is part of what
the construction costs.

Splits are by repository, never by claim. Claims mined from one commit
share a diff, and every arm sees that diff, so a random claim-level split
puts correlated items on both sides and reports memorization as
generalization. All prompt development used \texttt{boltons} and
\texttt{arrow}; every other repository is held out and was never inspected
while designing an arm. The held-out split contains 604 stratified claims
with 151 flips drawn from 17 repositories.

\begin{table*}[t]\centering
\caption{The mined corpus. Ten of the twenty-three repositories produced
any flips at all; the empty ones are listed because the rate at which
mining comes up dry is part of what the construction costs.}
\label{tab:corpus}
\begin{tabular}{lrrrrr}
\toprule
Repository & Commits & Claims & Flips & Flip rate & Discarded \\
\midrule
arrow & 26 & 1547 & 19 & 1.2\% & 73 \\
boltons & 6 & 360 & 14 & 3.9\% & 0 \\
cachetools & 18 & 1080 & 43 & 4.0\% & 180 \\
click & 18 & 1068 & 8 & 0.7\% & 0 \\
freezegun & 7 & 309 & 1 & 0.3\% & 0 \\
inflection & 2 & 120 & 0 & 0.0\% & 0 \\
itsdangerous & 7 & 420 & 2 & 0.5\% & 120 \\
jinja & 8 & 480 & 1 & 0.2\% & 0 \\
jsonschema & 5 & 297 & 0 & 0.0\% & 0 \\
markdown & 14 & 710 & 28 & 3.9\% & 0 \\
more-itertools & 16 & 960 & 41 & 4.3\% & 0 \\
packaging & 17 & 956 & 5 & 0.5\% & 0 \\
pluggy & 2 & 94 & 4 & 4.3\% & 0 \\
rich & 11 & 550 & 3 & 0.5\% & 0 \\
sortedcontainers & 4 & 240 & 0 & 0.0\% & 0 \\
structlog & 5 & 300 & 14 & 4.7\% & 0 \\
tenacity & 11 & 436 & 1 & 0.2\% & 0 \\
toolz & 7 & 262 & 0 & 0.0\% & 0 \\
typeguard & 3 & 180 & 0 & 0.0\% & 0 \\
\midrule
Total & 187 & 10369 & 184 & 1.8\% & 373 \\
\bottomrule
\end{tabular}

\end{table*}

\section{Arms}

\begin{itemize}
\item \textbf{A0, A1.} Always and never invalidate. A0 has recall 1 and precision
equal to the base rate.
\item \textbf{A3 file anchoring.} Invalidate if the claim's test imports any module
the diff touched. This is the content-digest class.
\item \textbf{A4 symbol anchoring.} Invalidate if a symbol the claim references had
its definition text changed. This is static change-impact analysis.
\item \textbf{A4b testmon.} \texttt{pytest-testmon}, a regression-test-selection
tool that records by coverage which source lines each test executes and
selects the affected tests after a change. Selection means invalidate.
This is dynamic change-impact analysis, and unlike A4 it is a tool
practitioners install rather than one we wrote. Note that it is not
comparable on cost: building its database requires executing the suite
under coverage instrumentation at the parent commit, which is the
per-commit execution cost the setting rules out.
\item \textbf{A5 semantic equivalence.} The model sees the diff and answers whether
it preserves behavior.
\item \textbf{A5C asymmetry control.} The model sees the claim \emph{and} the diff,
and still answers whether the diff preserves behavior. Same information as
A6, same question as A5.
\item \textbf{A6 claim-relative.} The model sees the claim and the diff and answers
whether that claim still holds.
\end{itemize}

\noindent\textbf{Strengthening the baselines.}
Beating a weak baseline shows nothing, so each was repaired until it was
fair. A4 needed three fixes. Keying definitions by bare name let unrelated
edits look like changes, because \texttt{pop} and \texttt{\_\_init\_\_}
recur across classes. It was blind to pytest fixtures, so a test written
as \texttt{def test\_x(self, cr)} never textually names the class under
test; this alone cost it all fourteen \texttt{structlog} flips. And
fixtures registered as \texttt{@pytest.fixture(name="cr")} do not match
their own function name. A5 was given a construction written against its
observed failure mode, which is over-firing, and told explicitly that
added functions, internal refactors, and unreachable paths preserve
behavior.

\section{Results}

\subsection{Held-out repositories}

Table~\ref{tab:arms} reports the held-out split: 244 claims, 61 flips,
stratified to a 25\% flip rate, with the natural rate recorded for
reweighting. All prompts were fixed before this split was evaluated.

\begin{table*}[t]\centering
\caption{Held-out split: 604 claims, 151 flips, 17 repositories,
stratified to a 25\% flip rate. Intervals are 95\% bootstrap.}
\label{tab:arms}
\begin{tabular}{llcc c}
\toprule
Arm & Signal & Precision & Recall & TP/FP/FN \\
\midrule
A3 file anchor & structural anchor & 0.21 [0.16,0.26] & 0.30 [0.23,0.38] & 46/173/105 \\
A4 symbol anchor & structural anchor & 0.39 [0.32,0.45] & 0.54 [0.46,0.62] & 81/128/70 \\
A4b testmon & structural anchor & 0.41 [0.36,0.47] & 0.87 [0.81,0.92] & 131/185/20 \\
A5 sem-equiv strict & diff-level LLM & 0.29 [0.24,0.34] & 0.69 [0.61,0.76] & 104/254/47 \\
A6 risk-framed & claim-level LLM & 0.79 [0.72,0.86] & 0.72 [0.64,0.79] & 108/28/43 \\
A4 $\cup$ A6 & union & 0.45 [0.39,0.51] & 0.81 [0.74,0.87] & 122/149/29 \\
\bottomrule
\end{tabular}

\end{table*}

\begin{figure*}[t]\centering
\includegraphics[width=0.82\textwidth]
  {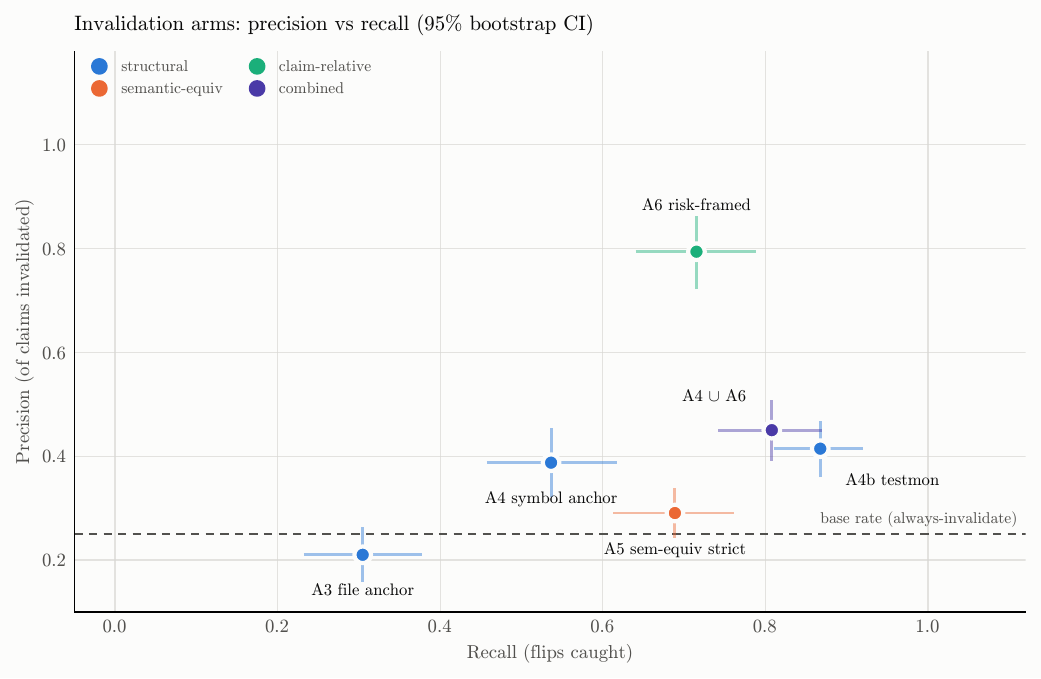}
\caption{Held-out arms with bootstrap 95\% confidence intervals. The
dashed line is the base rate, which is what always-invalidate achieves.
Both the file-anchoring and semantic-equivalence arms sit on or below it.}
\label{fig:pr}
\end{figure*}

A6 catches 108 of 151 flips against A4's 81, and does so while raising 28
false alarms to A4's 128. It is ahead on both axes at once, so no matched
operating point is needed to compare them. The precision confidence
intervals are disjoint, and a paired McNemar test on per-claim correctness
gives $p = 1\times10^{-20}$: A6 is correct where A4 is wrong on 162
claims, the reverse on 35.

Our pre-registered criterion required A6 to exceed A4 by at least 15
points of precision at matched recall. The measured gap is 40.7 points of
precision at 18 points more recall, which is strictly stronger than the
criterion asked for.

The semantic-equivalence arm behaves as it did on the smaller corpus. Even
in its strengthened form it fires on 59\% of held-out claims and reaches
0.291 precision against a 0.25 base rate, which is to say it is close to
invalidating everything.

\subsection{Against a real test-selection tool}

A4 is our own implementation, so we also ran \texttt{pytest-testmon},
which decides the same question with coverage-derived dependency data.

\begin{table*}[t]\centering
\caption{Held-out, 604 claims. Neither arm dominates: testmon buys low
staleness with a large verification budget, A6 buys a small budget at
higher staleness. testmon supplied a verdict for 80\% of claims; the rest
are charged to it as holds.}
\label{tab:testmon}
\begin{tabular}{lcccc}
\toprule
Arm & Precision & Recall & Claims re-verified & Flips served stale \\
\midrule
A4b testmon & 0.415 & \textbf{0.868} & 52\% & \textbf{13\%} \\
A6 risk-framed & \textbf{0.794} & 0.715 & \textbf{22\%} & 28\% \\
\bottomrule
\end{tabular}
\end{table*}

A6 does not dominate here, and we do not claim it does. The two arms sit
at different points on the cost/staleness frontier, and which one an
operator wants depends on what a stale answer costs relative to a
re-verification.

The informative number is testmon's precision. It catches 131 of the 151
flips, recall 0.868, and reaches 0.415 precision while doing so. Testmon
knows by direct coverage measurement which tests execute the changed
lines, so this is close to an upper bound on what dependency information
alone can deliver. It identifies what a change \emph{could} affect. Most
tests that a change could affect do not flip. That gap between impact and
invalidation is the effect this paper is about, and a tool with
essentially complete dependency knowledge does not close it.

A paired McNemar test still separates them clearly: A6 is correct where
testmon is wrong on 178 claims, the reverse on 44, $p = 3\times10^{-20}$.

The comparison is also asymmetric in testmon's favour on information and
against it on cost. Testmon requires a coverage-instrumented run of the
suite at the parent commit; A6 reads text and runs nothing.

\subsection{The question, not the evidence}

The obvious objection to the above is that A6 reads the claim and A5 does
not, so the gap could be an information asymmetry. A5C tests this by
giving the behavior-preservation judge the claim text.

\begin{table*}[t]\centering
\caption{Precision; the base rate is 0.25 in every column. The design
replicates on the post-cutoff split, whose commits postdate the training
data of every model evaluated, so the effect is not recall of the
repository.}
\label{tab:2x2}
\begin{tabular}{llcccc}
\toprule
& & \multicolumn{2}{c}{held-out} & \multicolumn{2}{c}{post-cutoff} \\
\cmidrule(lr){3-4}\cmidrule(l){5-6}
Information & Question & flash-lite & opus-5 & flash-lite & opus-5 \\
\midrule
diff only    & behavior-preserving?  & 0.291 & 0.309 & 0.338 & 0.278 \\
claim + diff & behavior-preserving?  & 0.301 & 0.325 & 0.349 & 0.289 \\
claim + diff & does this claim hold? & \textbf{0.794} & \textbf{0.974}
                                     & \textbf{0.750} & \textbf{0.913} \\
\midrule
\multicolumn{2}{l}{\emph{adding the claim is worth}}
  & +0.01 & +0.02 & +0.01 & +0.01 \\
\multicolumn{2}{l}{\emph{changing the question is worth}}
  & +0.49 & +0.65 & +0.40 & +0.62 \\
\bottomrule
\end{tabular}
\end{table*}

\begin{figure*}[t]\centering
\includegraphics[width=0.78\textwidth]
  {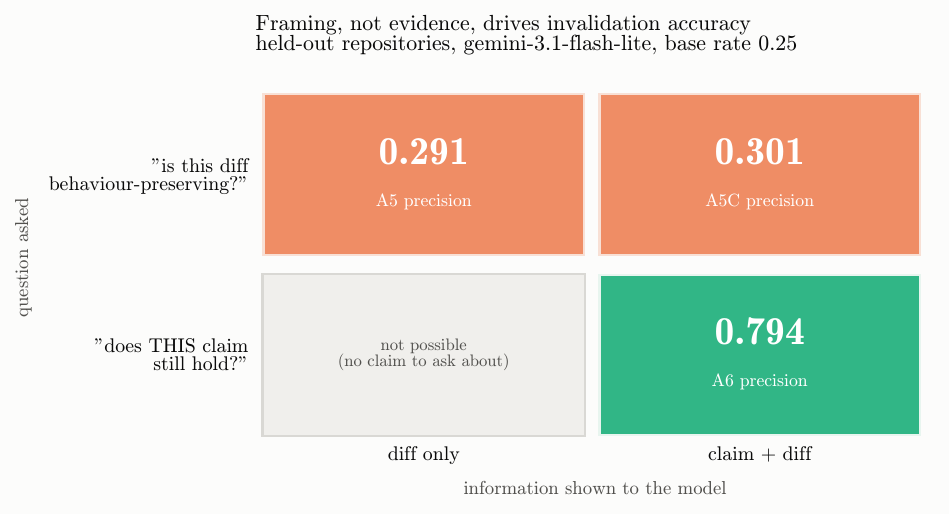}
\caption{The same data as Table~\ref{tab:2x2} for the budget model.
Reading across the top row, information buys almost nothing. Reading down
the right column, the question buys 0.59.}
\label{fig:design}
\end{figure*}

Paired McNemar for A6 against A5C on the budget model: A6 is correct where
A5C is wrong on 84 claims, the reverse on 9, $p = 2\times10^{-16}$.

Supplying the claim does move A5C's behavior, just not its accuracy. On the held-out split it fires on 70\% of claims for the
frontier model and reaches recall 0.972, so it invalidates nearly
everything and buys that recall by giving up all selectivity. That is the same degeneracy A5
shows without the claim. The judge reads the claim and still answers the
question it was asked.

\subsection{Is one repository carrying the result?}

Flips cluster. Behaviour-changing commits are not spread evenly, and most
repositories contribute none at all: of the 23 we mined, ten produced any
flips and two, \texttt{cachetools} and \texttt{more-itertools}, supply 84
of the 151 held-out positives between them. The obvious worry is that the
headline is a couple of repositories in disguise.

Table~\ref{tab:loo} drops each repository in turn and recomputes the gap
over all 17 repositories present in the held-out split. The smallest gap
across every exclusion is +34.4 points, which occurs when
\texttt{cachetools} is removed. Removing \texttt{more-itertools}, the
other large contributor, widens the gap to +46.7.

\begin{table*}[t]\centering
\caption{Leave-one-repository-out over the held-out split. Removing any
single repository, including either of the two largest contributors of
positives, leaves the gap far above the pre-registered threshold of +15.}
\label{tab:loo}
\begin{tabular}{lccc}
\toprule
Repository excluded & A6 precision & A4 precision & Gap \\
\midrule
none (all held-out) & 0.794 & 0.388 & +40.7 \\
\texttt{cachetools} (43 flips) & 0.775 & 0.431 & +34.4 \\
\texttt{more-itertools} (41 flips) & 0.705 & 0.238 & +46.7 \\
\texttt{markdown} (28 flips) & 0.793 & 0.382 & +41.1 \\
\texttt{structlog} (14 flips) & 0.788 & 0.345 & +44.3 \\
\multicolumn{4}{l}{\emph{\ldots\ 13 further repositories, gap between +37.7
and +43.4}} \\
\midrule
worst case over all 17 exclusions & n/a & n/a & \textbf{+34.4} \\
\bottomrule
\end{tabular}
\end{table*}

\subsection{What a practitioner buys}

Precision and recall on a stratified set do not tell an operator what to
budget. Fix an acceptable rate of serving a flipped claim, then ask how
much re-verification each arm demands to stay under it.
Table~\ref{tab:deploy} reports both that and precision reweighted to the
natural flip rate, which is what a deployment actually faces.

\begin{table*}[t]\centering
\caption{Held-out, 604 claims. The last column reweights held claims to
the observed 1.78\% natural flip rate, which is what a deployment faces.}
\label{tab:deploy}
\begin{tabular}{lccc}
\toprule
Arm & Claims re-verified & Flips served stale & Precision at natural rate \\
\midrule
A0 always-invalidate & 100\% & \textbf{0\%} & 0.018 \\
A3 file anchor & 36\% & 70\% & 0.014 \\
A4 symbol anchor & 35\% & 46\% & 0.033 \\
A4b testmon & 52\% & 13\% & 0.037 \\
A5 sem-equiv strict & 59\% & 31\% & 0.022 \\
A6 claim-relative & \textbf{22\%} & 28\% & \textbf{0.174} \\
\bottomrule
\end{tabular}
\end{table*}

Three readings matter. A6 serves fewer stale claims than symbol anchoring
(28\% against 46\%) while re-verifying under two thirds as much of the
store (22\% against 35\%), so it is not trading staleness for budget
against that baseline; it is better on both. Testmon reaches the lowest
staleness of any predictive arm, 13\%, but spends 52\% of the budget to do
it and must execute the suite first.

The last column is the one a practitioner should read. Under the natural
base rate every arm's precision collapses, because held claims outnumber
flipped ones by roughly fifty to one. A6 retains 0.174 against a 0.018
base rate, a tenfold lift. The impact-analysis arms reach 0.033 and 0.037,
under half the lift, and semantic equivalence reaches 0.022, which is
almost exactly what invalidating everything achieves.

\subsection{Do the results hold for prose claims?}
\label{sec:prose}

Our claims are test functions; a memory system stores sentences. To check
that the effect is about claims rather than about Python, we rewrote every
held-out claim as a prose note and re-ran the arm unchanged.

The rewriting is done in a separate pass that sees only the parent version
of the test. It has no access to the diff, the child commit, or the
oracle label, so it cannot encode the answer. It is instructed to state
what the library is asserted to do, including the names, arguments and
return shapes the claim depends on, and not to mention tests. A typical
output reads: \emph{``The \texttt{unique\_everseen} function filters an
iterable to return only the first occurrence of each element, preserving
the original order. It supports unhashable types like lists.''} The oracle
is untouched, so the numbers are directly comparable.

\begin{table*}[t]\centering
\caption{Held-out, 601 claims that rendered, 100\% coverage. Rewriting the
claim as prose costs precision but does not remove the effect.}
\label{tab:prose}
\begin{tabular}{lccc}
\toprule
Claim given to the arm & Precision & Recall & TP/FP/FN \\
\midrule
test function (as elsewhere) & 0.794 & 0.715 & 108/28/43 \\
prose note & 0.711 & 0.669 & 101/41/50 \\
\midrule
\emph{for reference: A4b testmon} & 0.415 & 0.868 & 131/185/20 \\
\emph{for reference: A5 sem-equiv} & 0.291 & 0.689 & 104/254/47 \\
\bottomrule
\end{tabular}
\end{table*}

Prose costs something real. Precision falls from 0.794 to 0.711 and a
paired test confirms the drop is not noise: the code form is correct where
prose is wrong on 46 claims, the reverse on 26, $p = 0.02$. That is the
expected direction. An assertion pins down exact values and call shapes,
and a sentence written from it keeps the behaviour but loses some of the
detail that makes a flip decidable.

What matters is that the effect survives the translation. At 0.711 the
prose arm is still far above every baseline measured on the same claims:
testmon at 0.415 ($p = 2\times10^{-13}$), symbol anchoring at 0.388
($p = 3\times10^{-16}$), and semantic equivalence at 0.291
($p = 1\times10^{-40}$). Whatever the claim-level question is buying, it
is not an artifact of the claim being executable Python.

\subsection{Model ladder}

\begin{table*}[t]\centering
\caption{Held-out, 604 claims, base rate 0.25. A5 spans 0.291 to 0.329
across a roughly 40$\times$ price range and never separates from the base
rate; A6 spans 0.705 to 0.974 on the same claims and diffs.}
\label{tab:ladder}
\begin{tabular}{llcccc}
\toprule
& & \multicolumn{2}{c}{A5 sem-equiv} & \multicolumn{2}{c}{A6 claim-relative} \\
\cmidrule(lr){3-4}\cmidrule(l){5-6}
Model & Tier & precision & fires on & precision & fires on \\
\midrule
gemini-3.1-flash-lite & floor   & 0.291 & 59\% & 0.794 & 23\% \\
gemini-2.5-flash      & floor   & 0.329 & 64\% & 0.810 & 23\% \\
claude-haiku-4.5      & mid     & 0.321 & 65\% & 0.705 & 29\% \\
claude-sonnet-5       & mid     & 0.322 & 63\% & 0.952 & 17\% \\
claude-opus-5         & ceiling & 0.309 & 72\% & \textbf{0.974} & 19\% \\
\bottomrule
\end{tabular}
\end{table*}

The degeneracy of A5 does not go away with scale. Across five models and
roughly a 40-fold range in price, its precision moves within a band of
0.04 and never leaves the neighbourhood of always-invalidating. The
frontier model is the worst of the five by this measure, not the best: it
fires on 72\% of claims. Asked the claim-level question on exactly the
same diffs, the same model reaches 0.974.

The cost frontier knees at the bottom. The cheapest model tested reaches
0.794 precision at \$0.025/M, within 0.18 of the frontier model at roughly
a fortieth of the price. This matters because an invalidation gate that
costs more than re-running the test has no reason to exist.

\subsection{Controls}

\noindent\textbf{Shuffled-diff null.} Each claim is paired with a diff from a
different commit, preferring a different repository. An arm that exploits
a genuine claim-diff dependence should stop firing. A6 goes from firing on
15\% of claims to 0\% at 100\% coverage. It never invalidates against an
unrelated diff.

\noindent\textbf{Memorization.} Every flip in the held-out split predates
2026-05, so all of them could be in pretraining data. We mined a separate
corpus restricted to commits after that date: 730 claims, 22 flips,
natural rate 3.0\%. On 88 stratified claims, A6 reaches 0.750 precision
with 0.955 recall on the budget model, and 0.913 precision with 1.000
recall on the frontier model. A5 stays at 0.338 and 0.278.

\noindent\textbf{Why the baselines reorder on that split.} The baselines
change places after the cutoff. A4 is stronger there (0.600 precision,
0.545 recall) while testmon is weaker (0.231, 0.545 at full coverage),
reversing the held-out ordering, and A6 beats both on precision and on
recall at the same time. Two things differ between the splits at once: the
commit dates, and the repositories. Table~\ref{tab:repocontrol} separates
them. The post-cutoff corpus draws on two repositories, and one of them,
\texttt{click}, also appears in the held-out split, so restricting both
splits to \texttt{click} holds the repository fixed and leaves the date as
the only difference.

Held that way, nothing moves. Every arm lands within 0.03 of its
cross-cutoff counterpart on both axes: A4 at 0.00/0.00 on either side,
testmon at 0.21/1.00 against 0.19/1.00, A6 at 0.53/1.00 against
0.50/1.00. The reordering comes from \texttt{boltons}, the repository
unique to the post-cutoff split, which has a 54\% flip rate and on which
A4 reaches 0.750 precision while testmon falls to 0.286. It is a
repository effect, not a recency effect, and the tradeoff against testmon
in Table~\ref{tab:testmon} is not contradicted by commits that postdate
the models' training data.

\begin{table*}[t]\centering
\caption{Holding the repository fixed across the training cutoff. Cells
are precision/recall for the budget model. \texttt{click} is the one
repository present on both sides, and every arm reproduces there to within
0.03. The reordering seen on the full post-cutoff split is carried by
\texttt{boltons}, which appears only after the cutoff and has a 54\% flip
rate.}
\label{tab:repocontrol}
\begin{tabular}{lcc cccc}
\toprule
Split & $n$ & flips & A4 & A4b & A5 & A6 \\
\midrule
post-cutoff, \texttt{click} & 62 & 8 & 0.00/0.00 & 0.21/1.00 & 0.20/1.00 & 0.53/1.00 \\
held-out, \texttt{click} & 68 & 8 & 0.00/0.00 & 0.19/1.00 & 0.19/1.00 & 0.50/1.00 \\
post-cutoff, \texttt{boltons} & 26 & 14 & 0.75/0.86 & 0.29/0.29 & 0.58/1.00 & 1.00/0.93 \\
post-cutoff, both repos & 88 & 22 & 0.60/0.55 & 0.23/0.55 & 0.34/1.00 & 0.75/0.95 \\
held-out, all 17 repos & 604 & 151 & 0.39/0.54 & 0.41/0.87 & 0.29/0.69 & 0.79/0.72 \\
\bottomrule
\end{tabular}

\end{table*}

With 22 positives the intervals are wide. We report this split as evidence
that the headline is not contamination, not as a second estimate.

\noindent\textbf{Paraphrase.} If A6 keys on the surface form of a claim rather
than on what it asserts, the interpretation of the 2$\times$2 weakens. We
rewrote 175 held-out claims syntactically: local bindings renamed to
meaningless identifiers, docstrings replaced, whitespace reflowed, with
attribute and call names left untouched so the assertions exercise exactly
the same API. Rewrites are produced by an AST pass and re-parsed, so no
model is involved. A6 agrees with its original verdict on 430 of 450
claims (95.6\%), and precision is 0.816 against 0.794 on the originals.
The residual disagreement is close to the run-to-run variance we measured
for an unchanged prompt.

\noindent\textbf{Rename class, and why it is absent.} Our second pre-registered
criterion was rename-class recall, on the reasoning that a pure rename is
the case where behavior-preservation must fail by construction. The class
turned out to be nearly empty: zero such claims in the development split
and one in the held-out split. Mature libraries with CI do not rename
public symbols casually; when they do, it is batched into a major release.
We report the criterion as not evaluable rather than as passed or failed
at $n=1$, and treat A5C as the mechanistic test instead, which isolates
the same effect without depending on one rare change class.

\section{Threats to validity}

\noindent\textbf{Construct validity: tests as claims.} Every claim we score is a
test function, because that is what admits a free and unambiguous oracle.
A memory system stores prose. If the effect we measure depended on claims
being executable Python, the paper would be about something narrower than
its title. Section~\ref{sec:prose} tests this directly by rewriting each
claim as a prose note, with the renderer given only the parent version of
the test and no access to the diff or the label, then re-running the arm
against the unchanged oracle. We still regard executable assertions as a
restricted sample of what a store holds: an assertion is exact and self-contained, where a note written by hand may
be vague or only partly checkable.
The benchmark therefore covers the checkable subset of claims, which is a
lower bound on claim diversity rather than a representative draw.

\noindent\textbf{Sample size.} 151 held-out positives and 22 post-cutoff
positives. Confidence intervals are reported throughout and several remain
wide, particularly on the post-cutoff split.

\noindent\textbf{Concentration of positives.} Flips are not spread evenly. Ten
of the 23 mined repositories produced any, and two supply 84 of the 151
held-out positives, so the effective sample is smaller than $n$ suggests.
The leave-one-repository-out analysis in Table~\ref{tab:loo} is the
control for this, and the conclusion survives every single-repository
deletion.

\noindent\textbf{Base rate.} The natural flip rate is 2.0\% even after targeted
mining. Precision reweighted to that rate is much lower than the
stratified numbers, and a deployment would face the reweighted figure.

\noindent\textbf{Scope.} Twenty-three Python libraries, none of them large, and
no other language. Two properties of this setting may not hold elsewhere.
Large codebases have longer dependency chains, which should make impact
analysis less precise still but also makes a claim harder to reason about
locally. Languages with static types may let a symbol-level baseline do
better than it does here. We make no claim about industrial monorepos or
about languages other than Python.

\noindent\textbf{Deployment.} No memory system was modified to use any of these
arms. The cost figures in Table~\ref{tab:deploy} follow from the confusion
matrices under a stated policy, not from an integration, so they describe
what an operator would buy rather than what one observed.

\noindent\textbf{Environment.} One shared virtual environment per repository
rather than per-commit pinning, which makes deep history (pre-Python-3
commits) unusable; those commits are excluded rather than mislabeled.

\noindent\textbf{Models.} Gemini and Claude only. Every OpenAI model identifier
on the gateway available to us returned 404.

\noindent\textbf{Prompt selection.} The risk-framed A6 prompt was chosen on the
development split. Its held-out and post-cutoff numbers are therefore the
honest ones and are what we report. All four prompts are reproduced in
full in Appendix~\ref{app:prompts}, so the question manipulation can be
read rather than taken on description.

\noindent\textbf{Measurement hazards.} Several intermediate results in this
project were artifacts, and a replication should guard against them. A
pytest collection error aborts an entire batch, and scoring a missing
verdict as a failure produced fifteen phantom flips from a commit that
touched an unrelated module. Walking history from \texttt{HEAD} rather
than an explicit branch silently changes the mined window after an
interrupted run. Tests inherited from mixin base classes have no source in
their own file, which left 31\% of claims with empty text. Passing a whole
suite as command-line arguments overflows the argument limit on large
repositories and yields nothing. A model that answers 58\% of claims and
scores 1.000 on them has not earned 1.000, so coverage must be reported.
Finally, on an early 31-flip pilot, A4 and A6 appeared to detect nearly
disjoint sets of flips (overlap 1 against 2.1 expected); on 61 held-out
flips the effect disappeared (18 against 14.8 expected). We report no
structural claim that has not survived the held-out split.

\section{Artifact}
\label{sec:artifact}

The benchmark, the mining pipeline, every arm, and the per-claim
predictions for each model and split are released. Two pieces are aimed at
reuse rather than at reproducing our tables.

\texttt{repro\_smoke.py} runs the whole pipeline on one repository and
three commits in under a minute and asserts the invariants that broke
silently during development: every claim carries its assertion text,
every retained claim passed 10 out of 10 runs at the parent, no partial
post-change verdict is kept, and the diffs handed to a model contain no
test files. It found a real defect the first time we ran it. Source
resolution for tests inherited from mixin base classes had been applied as
a post-hoc repair rather than inside the miner, so a fresh mine silently
produced claims with no text. Anyone extending this work will hit the same
class of problem, and the checks are cheaper than the debugging.

\texttt{test\_miner.py} pins the parsing behaviour that produced our worst
measurement error: a pytest batch aborted by a collection error must yield
no verdicts at all, rather than verdicts that read as failures.

\section{Related work}

\noindent\textbf{Memory and staleness.} EA-Graph anchors verification claims to
the content used to establish them and withdraws a claim when its support
changes~\cite{eagraph}. Work on temporal validity in retrieval memory
reports that embedding similarity separates a contradicted fact from a
duplicated one at close to chance, AUROC 0.59, because a contradiction is
often more similar to the original than a paraphrase
is~\cite{temporalvalidity}. PrecisionMemBench measures retrieval precision
independently of the generating model, on the argument that a system which
returns its whole belief store scores perfect recall while hiding severe
precision failures~\cite{precisionmembench}. These establish that precision
is the right metric to report. Their ground truth is annotation or a formal
oracle over synthetic data; ours is execution.

\noindent\textbf{Semantic change detection.} ChangeGuard reaches 77.1\%
precision and 69.5\% recall on semantics-changing changes using pairwise
learning-guided execution~\cite{changeguard}. SemaDiff identifies
semantic-changing commits with generated tests~\cite{semadiff}. These
answer whether behavior changed, which we use as the A5 arm and find to be
the wrong predicate for this task.

\noindent\textbf{Regression test selection and change impact analysis.} Deciding
which tests a change can affect has been studied for three
decades~\cite{rothermel96,rothermel97}, with surveys covering both
classical~\cite{rtssurvey} and learning-based~\cite{mltsp} selectors.
Ekstazi made the dynamic file-dependency approach practical~\cite{ekstazi},
and \texttt{pytest-testmon}~\cite{testmon}, our A4b arm, applies the same
idea to pytest. Our A4 is static symbol-level impact analysis.

We position against this literature deliberately, because the decision an
invalidation gate makes looks like the decision an RTS tool makes. The
results say they are not the same decision. A selector answers ``could
this change reach this test'', and on the claims testmon tracked it
answers that question perfectly, recall 1.000. It still reaches 0.412
precision, because most tests a change can reach do not flip. Safety, the
property RTS is designed for, is the wrong objective when the cost of a
false alarm is re-verification of a memory store rather than one extra
test execution.

The closest prior work is FineRTS~\cite{finerts}, which improves selection
precision by reasoning about whether a change modifies semantics rather
than treating any touched dependency as relevant. That is the same
observation one step removed: they refine which \emph{tests} to re-run by
asking about the change, while we ask about each stored claim directly.
Their target is a safe superset for a test runner; ours is a precision
gate for a memory store that cannot afford to run anything.

\section{Conclusion}

Invalidation accuracy in these experiments is determined by what the model
is asked, not by what it is shown or how large it is. A budget model asked
the claim-level question beats a frontier model asked the diff-level
question by a wide margin, and the frontier model asked the diff-level
question does no better than always invalidating.

The result we did not expect is that dependency information does not close
the gap either. A coverage-based test selector with perfect recall on the
claims it tracks still reaches 0.412 precision, because knowing which
tests a change can reach is not the same as knowing which claims it
falsifies. Impact analysis and invalidation are different predicates, and
the second one appears to require reasoning about the claim.

The practical reading is that a memory system should store claims in a
form it can interrogate one at a time, and should ask about them
individually rather than classifying the change. The benchmark, splits,
and per-arm predictions are released so the construction can be checked
and the numbers recomputed.

\appendices
\section{Prompts}
\label{app:prompts}

Every model-based arm in this paper is a single system prompt over the same
user message, so the comparisons stand or fall on the exact wording. All
four are reproduced here in full. The user message is the diff alone for
A5, and \texttt{CLAIM:\textbackslash n<test source>\textbackslash
n\textbackslash nDIFF:\textbackslash n<diff>} for A5C, A6 base and A6 risk,
with the diff truncated at 12{,}000 characters and test files excluded from
it by construction. No few-shot examples, no chain-of-thought instruction,
and no retries are used in any arm.

\subsection*{A5, semantic equivalence (strict)}

Diff only. One call per commit, its verdict reused across every claim of
that commit. \texttt{semantics\_preserving} true maps to \emph{held},
false to \emph{flipped}. This is the tightened construction: the earlier
version over-fired on added functions and internal refactors, so those
cases are ruled out by name rather than left to the model.

{\footnotesize\begin{verbatim}
You are judging whether a git diff changes externally
observable behavior for code that ALREADY existed
before the diff. Answer ONLY with
{"semantics_preserving": true or false}.
Say true (behavior-preserving) when the diff only: adds
new functions, classes, parameters with defaults, or
tests; refactors internals without changing outputs;
edits comments, docstrings, typing, or formatting; or
changes code paths no pre-existing caller reaches.
Say false ONLY when pre-existing callers would observe
a difference: changed return values or types, renamed
or removed public symbols, changed defaults or
signatures, changed exceptions, or changed ordering.
\end{verbatim}}

\subsection*{A5C, the information control}

Claim and diff, the same evidence A6 receives, but still asked A5's
question. The only edit against A5 is the second sentence, which
introduces the test and tells the model to judge the diff rather than the
test. Everything after that is A5 verbatim, which is what makes the
comparison a question manipulation rather than a prompt rewrite.

{\footnotesize\begin{verbatim}
You are judging whether a git diff changes externally
observable behavior for code that ALREADY existed
before the diff. You are also shown one test from the
codebase, purely as context for what the code is
expected to do. Judge the DIFF, not that test. Answer
ONLY with {"semantics_preserving": true or false}.
Say true (behavior-preserving) when the diff only: adds
new functions, classes, parameters with defaults, or
tests; refactors internals without changing outputs;
edits comments, docstrings, typing, or formatting; or
changes code paths no pre-existing caller reaches.
Say false ONLY when pre-existing callers would observe
a difference: changed return values or types, renamed
or removed public symbols, changed defaults or
signatures, changed exceptions, or changed ordering.
\end{verbatim}}

\subsection*{A6 base, the claim-relative question}

Claim and diff, asked whether this claim's assertions still pass. The
binary form, before risk framing.

{\footnotesize\begin{verbatim}
You will be shown a claim (a test function asserting
something about a codebase) and a git diff applied to
that codebase. Decide whether the claim's assertions
would still pass after the diff. Answer ONLY with a
JSON object of the form {"verdict": "held" or
"flipped"}. "held" = the assertions still pass.
"flipped" = the diff makes them fail.
\end{verbatim}}

\subsection*{A6 risk, the arm we report}

The same question under a recall-oriented instruction. Chosen on the
development split; every held-out and post-cutoff number in this paper
comes from this prompt, unchanged.

{\footnotesize\begin{verbatim}
You will be shown a claim (a test function asserting
something about a codebase) and a git diff applied to
that codebase. Your job is to catch claims that this
diff invalidates. Consider every value, signature,
default, name, exception, and ordering the claim
depends on. If there is ANY plausible way the diff
makes the claim's assertions fail, say flipped. Only
say held if the claim is clearly unaffected. Answer
ONLY with {"verdict": "held" or "flipped"}.
\end{verbatim}}

A5C is A5 with two sentences inserted and nothing else changed, which is
what makes it a control rather than a rewrite. A5C and A6 then receive the
same user message and return the same one-key JSON, and differ only in
what they are asked. That is the whole intervention behind the
2$\times$2 in Table~\ref{tab:2x2}, and precision moves by 40 to 67 points
across it.

\bibliographystyle{plain}
% Kept on one page by tightening list geometry rather than shrinking the
% type further, which was becoming hard to read.
% The references overrun the page by three entries; let this one page take
% the extra lines rather than shrinking the type or spilling a stub page.
\sloppy
{\small
\setlength{\itemsep}{0pt}

}

\end{document}